\documentclass[10pt,twocolumn,letterpaper]{article}

\newcommand\blfootnote[1]{%
\begingroup
\renewcommand\thefootnote{}\footnote{#1}%
\addtocounter{footnote}{-1}%
\endgroup
}

\usepackage{cvpr}
\usepackage{times}
\usepackage{graphicx}
\usepackage{multirow}
\usepackage{amsmath,amssymb} 
\usepackage{color}
\usepackage{caption}
\usepackage{subcaption}
\usepackage{booktabs}
\usepackage[pagebackref=true,breaklinks=true,letterpaper=true,colorlinks,bookmarks=false]{hyperref} 

\cvprfinalcopy 

\def\cvprPaperID{723} 

\ifcvprfinal\fi
\newcommand{\nrsfm}[0]{NRSfM }
\providecommand{\etal}{\textit{et al.}\xspace}

\begin{document}
\title{Category-Specific Object Reconstruction from a Single Image}
\author{Abhishek Kar$^*$, Shubham Tulsiani$^*$, Jo\~{a}o Carreira and Jitendra Malik\\
University of California, Berkeley - Berkeley, CA 94720\\
{\tt\small \{akar,shubhtuls,carreira,malik\}@eecs.berkeley.edu}}

\maketitle
\thispagestyle{empty}
\blfootnote{* Authors contributed equally}
\begin{abstract}
Object reconstruction from a single image -- in the wild -- is a problem where we can make progress and get meaningful results today. This is the main message of this paper, which introduces an automated pipeline with pixels as inputs and 3D surfaces of various rigid categories as outputs in images of realistic scenes. At the core of our approach are deformable 3D models that can be learned from 2D annotations available in existing object detection datasets, that can be driven by noisy automatic object segmentations and which we complement with a bottom-up module for recovering high-frequency shape details. We perform a comprehensive quantitative analysis and ablation study of our approach using the recently introduced PASCAL 3D+ dataset and show very encouraging automatic reconstructions on PASCAL VOC.




\end{abstract}

\section{Introduction}
\begin{figure}[htb!]
\includegraphics[width = 0.48\textwidth]{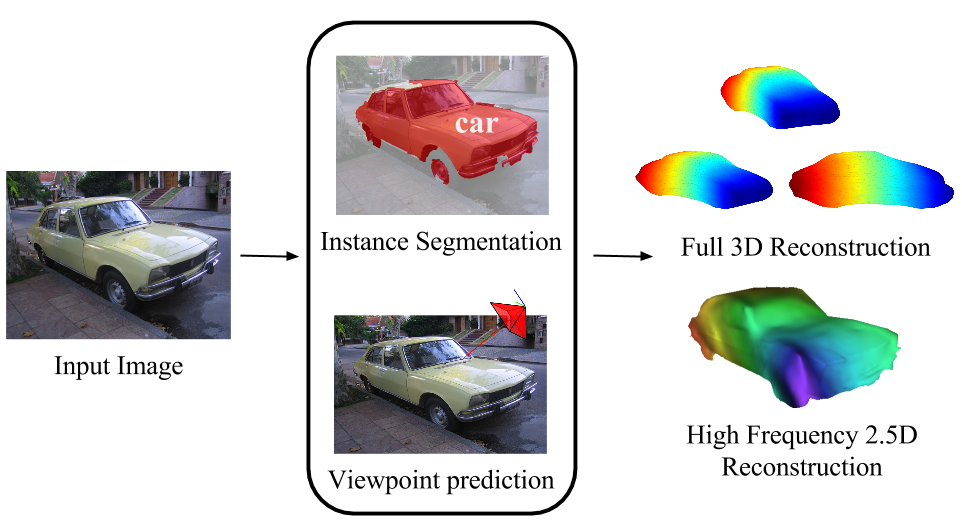}
\caption{Automatic object reconstruction from a single image obtained by our system. Our method leverages estimated instance segmentations and predicted viewpoints to generate a full 3D mesh and high frequency 2.5D depth maps.}
  \label{fig:fig1}
\end{figure}

Consider the car in Figure \ref{fig:fig1}. As humans, not only can we infer at a glance that the image contains a car, we also construct a rich internal representation of it such as its location and 3D pose. Moreover, we have a guess of  its 3D shape, even though we might never have have seen this particular car. We can do this because we don't experience the image of this car {\em tabula rasa}, but in the context of our  ``remembrance of things past".   Previously seen cars enable us to develop a notion of the 3D shape of cars, which we can project to this particular instance. We also specialize our representation to this particular instance (e.g. any custom decorations it might have), signalling that both top-down and bottom-up cues influence our percept~\cite{nandakumar2011little}.

A key component in such a process would be a mechanism to build  3D shape models from past visual experiences. We have developed an algorithm that can build category-specific shape models from just images with 2D annotations (segmentation masks and a small set of keypoints) present in modern computer vision datasets (e.g. PASCAL VOC~\cite{pascal-voc-2012}).  These models are then used to guide the top down 3D shape reconstruction of novel 2D car images. We complement our top-down shape inference algorithm with a bottom-up module that further refines our shape estimate for a particular instance. Finally, building upon the rapid recent progress in recognition modules~\cite{arbelaez2014multiscale,carreira_cvpr10,girshick2013rich,BharathECCV2014,ShubhamPose} (object detection, segmentation and pose estimation), we demonstrate that our learnt models are robust when applied ``in the wild" enabling fully automatic reconstructions with just images as inputs.

\begin{figure*}[htb!]
\includegraphics[width = \textwidth]{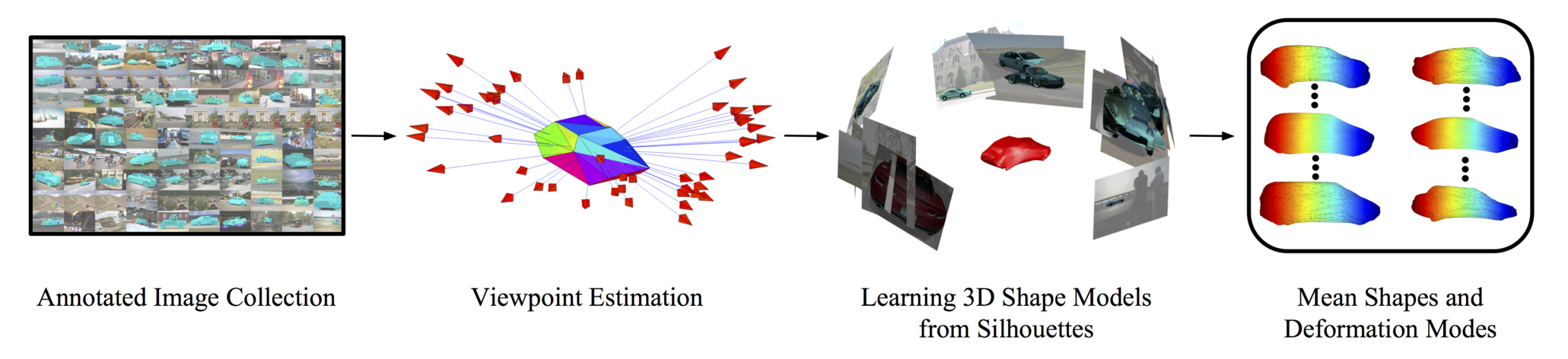}
\caption{Overview of our training pipeline. We use an annotated image collection to estimate camera viewpoints which we then use alongwith object silhouettes to learn 3D shape models. Our learnt shape models, as illustrated in the rightmost figure are capable of deforming to capture intra-class shape variation.}
\label{fig:figtrain}
\end{figure*}

The recent method of Vicente \etal~\cite{carvi14} reconstructs 3D models from similar annotations as we do but it has a different focus: it aims to reconstruct a fully annotated image set while making strong assumptions about the quality of the segmentations it fits to and is hence inappropriate for reconstruction in an unconstrained setting. Our approach can work in such settings, partly because it uses  explicit 3D shape models. Our work also has connections to that of Kemelmacher-Shlizerman \etal~\cite{kemelmacher2011internet,movingFace:eccv14} which aims to learn morphable models for faces from 2D images, but we focus on richer shapes in unconstrained settings, at the expense of lower resolution reconstructions.   

In the history of computer vision,  model-based object reconstruction from a single image has reflected varying preferences on model representations.  Generalized cylinders \cite{nevatia1977description} resulted in very compact descriptions for certain classes of shapes, and can be used for category level descriptions, but the fitting problem for general shapes in challenging. Polyhedral models \cite{gupta2010blocks,xiao2012localizing}, which trace back to the early work of Roberts \cite{roberts1963machine}, and CAD models \cite{limparsing,satkin20143dnn} provide crude approximations of shape and given a set of point correspondences can be quite effective for determining instance viewpoints.  Here we pursue more expressive basis shape models \cite{Anguelov:SCAPE2005,blanz1999morphable,zia2013detailed} which establish a balance between the two extremes as they can deform but only along class-specific modes of variation. In contrast to previous work (e.g. \cite{zia2013detailed}), we fit them to automatic figure-ground object segmentations.

Our paper is organized as follows: in Section \ref{sec:learning} we describe our model learning pipeline where we estimate camera viewpoints for all training objects (Section \ref{sec:nrsfm}) followed by our shape model formulation (Section \ref{sec:basisshapes}) to learn 3D models. Section \ref{sec:testing} describes our testing pipeline where we use our learnt models to reconstruct novel instances without assuming any annotations. We evaluate our reconstructions under various settings in Section \ref{sec:experiments} and provide sample reconstructions in the wild.

\section{Learning Deformable 3D Models}
\label{sec:learning}
We are interested in 3D shape models that can be robustly aligned to noisy object segmentations by incorporating top-down class-specific knowledge of how shapes from the class typically project into the image. We want to learn such models from just 2D training images, aided by ground truth segmentations and a few keypoints, similar to \cite{carvi14}. Our approach operates by first estimating the viewpoints of all objects in a class using a structure-from-motion approach, followed by optimizing over a deformation basis of representative 3D shapes that best explain all silhouettes, conditioned on the viewpoints. We describe these two stages of model learning in the following subsections. Figure \ref{fig:figtrain} illustrates this training pipeline of ours.

\subsection{Viewpoint Estimation}
\label{sec:nrsfm}
We use the framework of NRSfM \cite{Bregler2000} to jointly estimate the camera viewpoints (rotation, translation and scale) for all training instances in each class. Originally proposed for recovering shape and deformations from video \cite{nrsfm_priors,Torresani2008NRSFM,varnrsfm2013,Bregler2000}, NRSfM is a natural choice for viewpoint estimation from sparse correspondences as intra-class variation may become a confounding factor if not modeled explicitly. However, the performance of such algorithms has only been explored on simple categories, such as SUV's \cite{Zhu_ModelEvolution:2010} or flower petal and clown fish \cite{prasad2010finding}. Closer to our work, Hejrati and Ramanan \cite{HejratiR12} used NRSfM on a larger class (cars) but need a predictive detector to fill-in missing data (occluded keypoints) which we do not assume to have here.


We closely follow the EM-PPCA formulation of Torresani \etal\cite{Torresani2008NRSFM} and propose a simple extension to the algorithm that incorporates silhouette information in addition to keypoint correspondences to robustly recover cameras and shape bases. Energies similar to ours have been proposed in the shape-from-silhouette literature \cite{balloonshapes} and with rigid structure-from-motion \cite{carvi14} but, to the best of our knowledge, not in conjunction with NRSfM.

\vspace{3mm}
\noindent \textbf{NRSfM Model.} Given $K$ keypoint correspondences per instance $n \in \{1,\cdots,N\}$, our adaptation of the NRSfM algorithm in \cite{Torresani2008NRSFM} corresponds to maximizing the likelihood of the following model:

\begin{equation}
\begin{aligned}
&{P_{n}} = (I_K\otimes c_nR_n)S_{n}+T_{n} + N_{n} \\
&S_n = \bar{S} + Vz_n \\
&z_n \sim \mathcal{N}(0,I), \quad N_{n}\sim \mathcal{N}(0,\sigma^2 I)
\end{aligned}
\end{equation}
\begin{align}
\text{subject to:}\nonumber &\quad R_nR_{n}^{T} = I_2 \\
\label{eq:sil_constraint}
\sum\limits_{k=1}^K C_{n}^{mask}(p_{k,n}) &= 0, \quad \forall n\in \{1,\cdots,N\}
\end{align}
Here, $P_n$ is the 2D projection of the 3D shape $S_n$ with white noise $N_n$ and the rigid transformation given by the orthographic projection matrix $R_n$, scale $c_n$ and 2D translation $T_n$. The shape is parameterized as a factored Gaussian with a mean shape $\bar{S}$, $m$ basis vectors $[V_1,V_2, \cdots,V_m] = V$ and latent deformation parameters $z_n$. Our key modification is constraint \eqref{eq:sil_constraint} where $C_{n}^{mask}$ denotes the Chamfer distance field of the $n^{th}$ instance's binary mask and says that all keypoints $p_{k,n}$ of instance $n$ should lie inside its binary mask. We observed that this results in more accurate viewpoints as well as more meaningful shape bases learnt from the data.

\begin{figure}[htb!]
  \centering
\includegraphics[width=.8\linewidth]{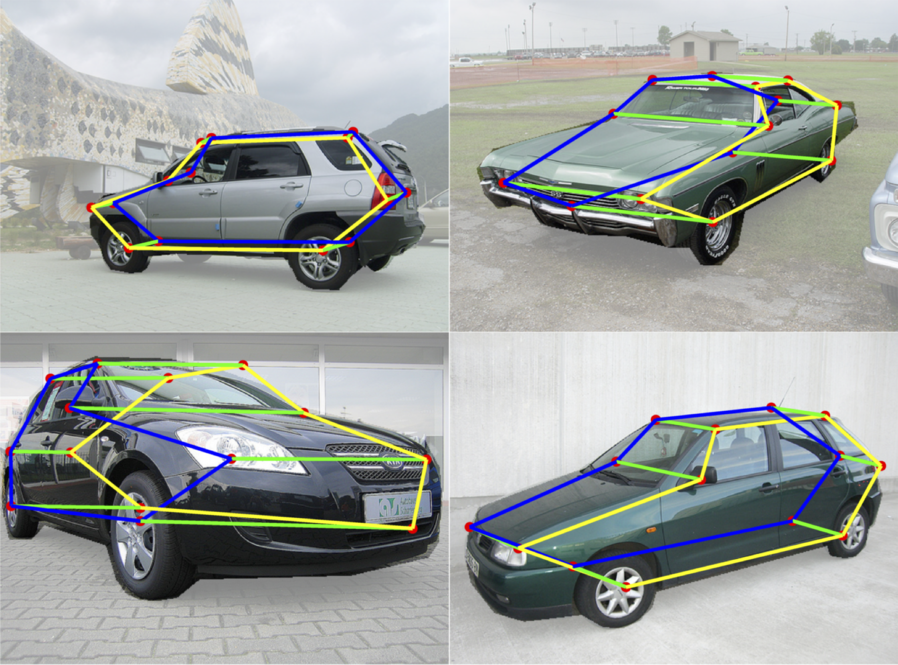}
\caption{NRSfM viewpoint estimation: Estimated viewpoints visualized using a 3D car wireframe.}
\label{fig:nrsfm_images}
\end{figure}

\vspace{3mm}
\noindent \textbf{Learning.} The likelihood of the above model is maximized using the EM algorithm. Missing data (occluded keypoints) is dealt with by ``filling-in" the values using the forward equations after the E-step. The algorithm computes shape parameters $\{\bar{S},V\}$, rigid body transformations $\{c_n,R_n,T_n\}$ as well as the deformation parameters $\{z_n\}$ for each training instance $n$. In practice, we augment the data using horizontally mirrored images to exploit bilateral symmetry in the object classes considered. We also precompute the Chamfer distance fields for the whole set to speed up computation. As shown in Figure \ref{fig:nrsfm_images}, \nrsfm  allows us to reliably predict viewpoint while being robust to intraclass variations.

\subsection{3D Basis Shape Model Learning}
\label{sec:basisshapes}
Equipped with camera projection parameters and keypoint correspondences (lifted to 3D by NRSfM) on the whole training set, we proceed to build deformable 3D shape models from object silhouettes within a class. 3D shape reconstruction from multiple silhouettes projected from a single object in calibrated settings has been widely studied. Two prominent approaches are \textit{visual hulls} \cite{laurentini1994hull} and variational methods derived from \textit{snakes} e.g \cite{esteban2004snake,yusuf2006snake} which deform a surface mesh iteratively until convergence. Some interesting recent papers have extended variational approaches to handle categories \cite{cashman2013dolphins,chen20123d} but typically require some form of 3D annotations to bootstrap models. A recently proposed visual-hull based approach \cite{carvi14} requires only 2D annotations as we do for class-based reconstruction and it was successfully demonstrated on PASCAL VOC but does not serve our purposes as it makes strong assumptions about the accuracy of the segmentation and will in fact fill entirely any segmentation with a voxel layer. 


\label{OPTIMIZ}
\vspace{3mm}
\noindent \textbf{Shape Model Formulation.} We model our category shapes as deformable point clouds -- one for each subcategory of the class. The underlying intuition is the following: some types of shape variation may be well explained by a parametric model e.g. a Toyota sedan and a Lexus sedan, but it is unreasonable to expect them to model the variations between sail boats and cruise liners. Such models typically require knowledge of object parts, their spatial arrangements etc.~\cite{Kalogerakis:2012:ShapeSynthesis} and involve complicated formulations that are difficult to optimize. We instead train separate linear shape models for different subcategories of a class. As in the \nrsfm model, we use a linear combination of bases to model these deformations. Note that we learn such models from silhouettes and this is what enables us to learn deformable models without relying on point correspondences between scanned 3D exemplars~\cite{blanz2003face}.


Our shape model $M = (\overline{S},V)$ comprises of a mean shape $\overline{S}$ and deformation bases $V = \{ V_1,.,V_K \} $ learnt from a training set $T:\{(O_i,P_i)\}_{i=1}^N$, where $O_i$ is the instance silhouette and $P_i$ is the projection function from world to image coordinates. Note that the $P_i$ we obtain using NRSfM corresponds to orthographic projection but our algorithm could handle perspective projection as well.

\vspace{3mm}
\noindent \textbf{Energy Formulation.} We formulate our objective function primarily based on image silhouettes. For example, the shape for an instance should always project within its silhouette and should agree with the keypoints (lifted to 3D by \nrsfm). We capture these by defining corresponding energy terms as follows: (here $P(S)$ corresponds to the 2D projection of shape $S$, $C^{mask}$ refers to the Chamfer distance field of the binary mask of silhouette $O$ and $\Delta^k(p;Q)$ is defined as the squared average distance of point $p$ to its $k$ nearest neighbors in set $Q$)

\vspace{3mm}
\noindent \textbf{Silhouette Consistency.} Silhouette consistency simply enforces the predicted shape for an instance to project inside its silhouette. This can be achieved by penalizing the points projected outside the instance mask by their distance from the silhouette. In our $\Delta$ notation it can be written as follows:
\begin{gather}
 \label{eq:sil_con}E_{s}(S,O,P)=\underset{C^{mask}(p)>0}{\sum}\Delta^1(p;O)
 \end{gather}

\vspace{3mm}
\noindent \textbf{Silhouette Coverage.}
Using silhouette consistency alone would just drive points projected outside in towards the silhouette. This wouldn't ensure though that the object silhouette is ``filled'' - i.e. there might be overcarving. We deal with it by having an energy term that encourages points on the silhouette to pull nearby projected points towards them. Formally, this can be expressed as:
\begin{gather}
    \label{eq:sil_cov}E_{c}(S,O,P)=\underset{p\in O}{\sum}\Delta^m(p;P(S))
\end{gather}


\vspace{3mm}
\noindent \textbf{Keypoint Consistency.} Our \nrsfm algorithm provides us with sparse 3D keypoints along with camera viewpoints. We use these sparse correspondences on the training set to deform the shape to explain these 3D points. The corresponding energy term penalizes deviation of the shape from the 3D keypoints $KP$ for each instance. Specifically, this can be written as:
\begin{gather}
    \label{eq:kpgrad}E_{kp}(S,O,P)=\underset{\kappa\in KP}{\sum}\Delta^m(\kappa;S)
\end{gather}

\vspace{3mm}
\noindent \textbf{Local Consistency.} In addition to the above data terms, we use a simple shape regularizer to restrict arbitrary deformations by imposing a quadratic deformation penalty between every point and its neighbors. We also impose a similar penalty on deformations to ensure local smoothness. The $\delta$ parameter represents the mean squared displacement between neighboring points and it encourages all faces to have similar size. Here $V_{ki}$ is the $i^{th}$ point in the $k^{th}$ basis.
\begin{multline}
 \label{eq:local_con}E_{l}(\bar{S},V)=\underset{i}{\sum}\underset{j\in N(i)}{\sum}((\|\bar{S}_{i}-\bar{S}_{j}\| - \delta)^2 +\\
 \underset{k}{\sum}\|V_{ki}-V_{kj}\|^2)
 \end{multline}
 
\vspace{3mm}
\noindent \textbf{Normal Smoothness.} Shapes occurring in the natural world tend to be locally smooth. We capture this prior on shapes by placing a cost on the variation of normal directions in a local neighborhood in the shape. Our normal smoothness energy is formulated as
\begin{gather}
 \label{eq:normal_con}E_{n}(S)=\underset{i}{\sum}\underset{j\in N(i)}{\sum}(1-\vec{\mathcal{N}_i} \cdot \vec{\mathcal{N}_j})
\end{gather}
 Here, $\vec{\mathcal{N}_i}$ represents the normal for the $i^{th}$ point in shape $S$ which is computed by fitting planes to local point neighborhoods. Our prior essentially states that local point neighborhoods should be flat. Note that this, in conjunction with our previous energies automatically enforces the commonly used prior that normals should be perpendicular to the viewing direction at the occluding contour~\cite{Barron2012B}.

Our total energy is given in equation \ref{eq:formulation}. In addition to the above smoothness priors we also penalize the $L_2$ norm of the deformation parameters $\alpha_i$ to prevent unnaturally large deformations.
\begin{multline}
\label{eq:formulation}
E_{tot}(\bar{S},V,\alpha) = E_{l}(\bar{S},V)+\\
\underset{i}{\sum}(E_{s}^i+E_{kp}^i+E_{c}^i+E_{n}^i+\underset{k}{\sum}(\|\alpha_{ik}V_k\|_F^{2}))
\end{multline}

\begin{figure}[htb!, width=0.45\textwidth]
\centering
  \includegraphics[width=\linewidth]{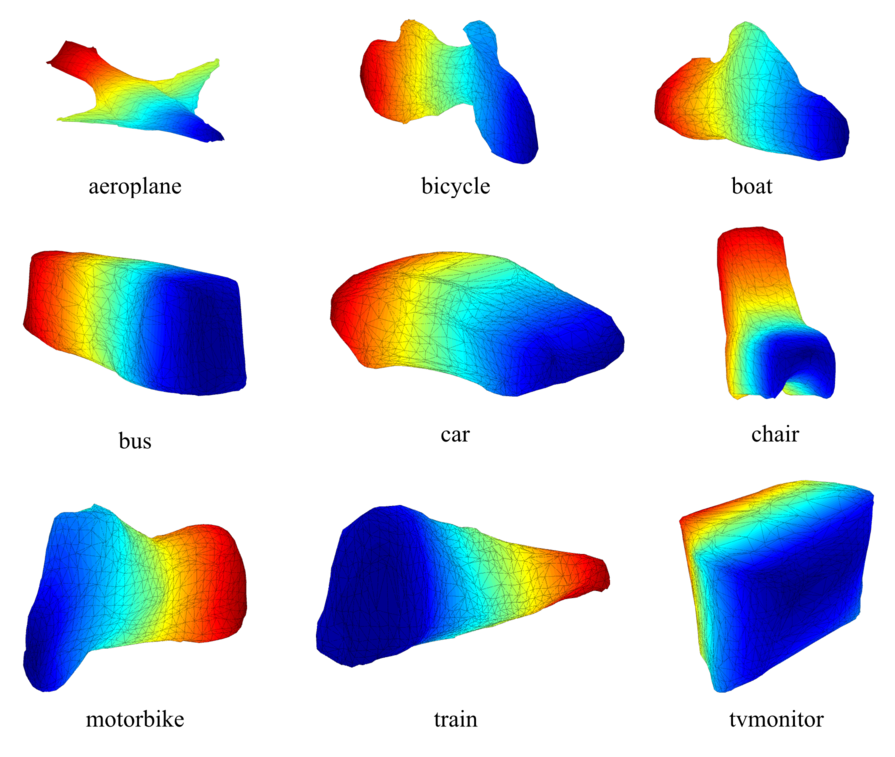}
  \caption{Mean shapes learnt for rigid classes in PASCAL VOC obtained using our basis shape formulation. Color encodes depth when viewed frontally.}
  \label{fig:meanDense}
\end{figure}

\vspace{3mm}
\noindent \textbf{Learning.} We solve the optimization problem in equation~\ref{eq:optimization} to obtain our shape model $M=(\bar{S},V)$. The mean shape and deformation basis are inferred via block-coordinate descent on $(\bar{S},V)$ and $\alpha$ using sub-gradient computations over the training set. We restrict $\|V_k\|_F$ to be a constant to address the scale ambiguity between $V$ and $\alpha$ in our formulation. In order to deal with imperfect segmentations and wrongly estimated keypoints, we use truncated versions of the above energies that reduce the impact of outliers. The mean shapes learnt using our algorithm for 9 rigid categories in PASCAL VOC are shown in Figure \ref{fig:meanDense}. Note that in addition to representing the coarse shape details of a category, the model also learns finer structures like chair legs and bicycle handles, which become more prominent with deformations.

\begin{equation}
\begin{aligned}
\label{eq:optimization}
& \underset{\bar{S},V,\alpha}{\text{min}}
& & E_{tot}(\bar{S},V,\alpha) \\
& \text{subject to:}
& & S^i = \bar{S} + \underset{k}{\sum}\alpha_{ik} V_k
\end{aligned}
\end{equation}

Our training objective is highly non-convex and  non-smooth and is susceptible to initialization. We follow the suggestion of \cite{esteban2004snake} and initialize our mean shape with a soft visual hull computed using all training instances. The deformation bases and deformation weights are initialized randomly. 

\section{Reconstruction in the Wild}
\label{sec:testing}
We approach object reconstruction from the big picture downward - like a sculptor first hammering out the big chunks and then chiseling out the details. After detecting and segmenting objects in the scene, we infer their coarse 3D poses and use them to fit our top-down shape models to the noisy segmentation masks. Finally, we recover high frequency shape details from shading cues. We will now explain these components one at a time.

%

\vspace{3mm}
\noindent \textbf{Initialization.} During inference, we first detect and segment the object in the image~\cite{BharathECCV2014} and then predict viewpoint (rotation matrix) and subcategory for the object using a CNN based system similar to \cite{ShubhamPose} (augmented to predict subcategories). Our learnt models are at a canonical bounding box scale - all objects are first resized to a particular width during training. Given the predicted bounding box, we scale the learnt mean shape of the predicted subcategory accordingly. Finally, the mean shape is rotated as per the predicted viewpoint and translated to the center of the predicted bounding box.

\vspace{3mm}
\noindent \textbf{Shape Inference.} After initialization, we solve for the deformation weights $\alpha$(initialized to $0$) as well as all the camera projection parameters (scale, translation and rotation) by optimizing equation~\eqref{eq:optimization} for fixed $\bar{S},V$.  Note that we do not have access to annotated keypoint locations at test time, the `Keypoint Consistency' energy $E_{kp}$ is ignored during the optimization.

\vspace{3mm}
\noindent \textbf{Bottom-up Shape Refinement.} \label{sec:sirfs} The above optimization results in a top-down 3D reconstruction based on the category-level models, inferred object silhouette, viewpoint and our shape priors. We propose an additional processing step to recover high frequency shape information by adapting the intrinsic images algorithm of Barron and Malik \cite{barronPAMI13,Barron2012B}, SIRFS, which exploits statistical regularities between shapes, reflectance and illumination 
Formally, SIRFS is formulated as the following optimization problem:
\begin{equation*}
\begin{aligned}
& \underset{Z,L}{\text{minimize}}
& & g(I-S(Z,L))+f(Z)+h(L)
\end{aligned}
\end{equation*}
where $R=I-S(Z,L)$ is a log-reflectance image, $Z$ is a depth map and $L$ is a spherical-harmonic model of illumination. $S(Z,L)$ is a rendering engine which produces a log shading image with the illumination $L$. $g$, $f$ and $h$ are the loss functions corresponding to reflectance, shape and illumination respectively.

We incorporate our current coarse estimate of shape into SIRFS through an additional loss term:

\begin{equation*}
f_o(Z,Z') = \underset{i}{\sum}((Z_i-Z_i')^2+\epsilon^2)^{\gamma_o}
\end{equation*}
where $Z'$ is the initial coarse shape and $\epsilon$ a parameter added to make the loss differentiable everywhere. We obtain $Z'$ for an object by rendering a depth map of our fitted 3D shape model which guides the optimization of this highly non-convex cost function. The outputs from this bottom-up refinement are reflectance, shape and illumination maps of which we retain the shape.

\vspace{3mm}
\noindent \textbf{Implementation Details.}
The gradients involved in our optimization for shape and projection parameters are extremely efficient to compute. We use approximate nearest neighbors computed using k-d tree to implement the `Silhouette Coverage' gradients and leverage Chamfer distance fields for obtaining `Silhouette Consistency' gradients. Our overall computation takes only about 2 sec to reconstruct a novel instance using a single CPU core. Our training pipeline is also equally efficient - taking only a few minutes to learn a shape model for a given object category.

\section{Experiments}
\label{sec:experiments}
\begin{table*}[htb!]
\centering
\begin{tabular}{|c|c|cccccccccc|c|}
\hline 
 & \multicolumn{1}{c}{\textbf{\footnotesize{}Classes}} & \textbf{\footnotesize{}aero} & \textbf{\footnotesize{}bike} & \textbf{\footnotesize{}boat} & \textbf{\footnotesize{}bus} & \textbf{\footnotesize{}car} & \textbf{\footnotesize{}chair} & \textbf{\footnotesize{}mbike} & \textbf{\footnotesize{}sofa} & \textbf{\footnotesize{}train} & \textbf{\footnotesize{}tv} & \textbf{\footnotesize{}mean}\tabularnewline
\hline 
\hline 
\multirow{3}{*}{\textbf{\footnotesize{}Mesh}} & {\footnotesize{}KP+Mask} & \textbf{\footnotesize{}5.00} & {\footnotesize{}6.27} & {\footnotesize{}9.94} & \textbf{\footnotesize{}6.22} & {\footnotesize{}5.18} & \textbf{\footnotesize{}5.20} & {\footnotesize{}4.98} & {\footnotesize{}6.58} & \textbf{\footnotesize{}12.60} & {\footnotesize{}9.64} & \textbf{\footnotesize{}7.16}\tabularnewline
 & {\footnotesize{}Carvi\cite{carvi14}} & {\footnotesize{}5.07} & \textbf{\footnotesize{}6.03} & \textbf{\footnotesize{}8.80} & {\footnotesize{}8.76} & \textbf{\footnotesize{}4.38} & {\footnotesize{}5.74} & \textbf{\footnotesize{}4.86} & \textbf{\footnotesize{}6.49} & {\footnotesize{}17.52} & \textbf{\footnotesize{}8.37} & {\footnotesize{}7.60}\tabularnewline
 & {\footnotesize{}Puffball\cite{twarog2012playing}} & {\footnotesize{}9.73} & {\footnotesize{}10.39} & {\footnotesize{}11.68} & {\footnotesize{}15.40} & {\footnotesize{}11.77} & {\footnotesize{}8.58} & {\footnotesize{}8.99} & {\footnotesize{}8.62} & {\footnotesize{}23.68} & {\footnotesize{}9.45} & {\footnotesize{}11.83}\tabularnewline
\hline 
\multirow{3}{*}{\textbf{\footnotesize{}Depth}} & {\footnotesize{}KP+Mask} & \textbf{\footnotesize{}9.25} & {\footnotesize{}7.87} & {\footnotesize{}12.36} & \textbf{\footnotesize{}11.77} & {\footnotesize{}7.22} & {\footnotesize{}7.51} & {\footnotesize{}8.97} & \textbf{\footnotesize{}9.70} & {\footnotesize{}30.91} & {\footnotesize{}6.84} & \textbf{\footnotesize{}11.24}\tabularnewline
 & {\footnotesize{}Carvi\cite{carvi14}} & {\footnotesize{}9.39} & \textbf{\footnotesize{}7.24} & \textbf{\footnotesize{}11.43} & {\footnotesize{}18.42} & \textbf{\footnotesize{}6.86} & \textbf{\footnotesize{}7.39} & \textbf{\footnotesize{}8.06} & {\footnotesize{}12.21} & \textbf{\footnotesize{}29.57} & \textbf{\footnotesize{}5.75} & {\footnotesize{}11.63}\tabularnewline
 & {\footnotesize{}SIRFS\cite{Barron2012B}} & {\footnotesize{}12.98} & {\footnotesize{}12.31} & {\footnotesize{}16.03} & {\footnotesize{}29.21} & {\footnotesize{}21.58} & {\footnotesize{}15.53} & {\footnotesize{}16.30} & {\footnotesize{}18.08} & {\footnotesize{}38.54} & {\footnotesize{}21.36} & {\footnotesize{}20.19}\tabularnewline
\hline 
\end{tabular}
\caption{\label{tab:carvi_compare} Studying the expressiveness of our learnt 3D models: comparison between our method and \cite{carvi14,twarog2012playing} using ground truth keypoints and masks on PASCAL VOC. Note that \cite{carvi14} operates with ground truth annotations and reconstructs an image corpus and our method is used here on the same task for a fair comparison. Please see text for more details.}
\end{table*}

\begin{table*}[htb!]
  \centering

\begin{tabular}{|c|c|cccccccccc|c|}
\hline 
 & \textbf{\footnotesize{}Classes} & \textbf{\footnotesize{}aero} & \textbf{\footnotesize{}bike} & \textbf{\footnotesize{}boat} & \textbf{\footnotesize{}bus} & \textbf{\footnotesize{}car} & \textbf{\footnotesize{}chair} & \textbf{\footnotesize{}mbike} & \textbf{\footnotesize{}sofa} & \textbf{\footnotesize{}train} & \textbf{\footnotesize{}tv} & \textbf{\footnotesize{}mean}\tabularnewline
\hline 
\hline 
\multirow{4}{*}{\textbf{\footnotesize{}Mesh}} & {\footnotesize{}KP+Mask} & {\footnotesize{}5.13} & {\footnotesize{}6.46} & {\footnotesize{}10.46} & {\footnotesize{}5.89} & {\footnotesize{}5.07} & {\footnotesize{}5.34} & {\footnotesize{}5.15} & {\footnotesize{}15.07} & {\footnotesize{}12.16} & {\footnotesize{}11.69} & {\footnotesize{}8.24}\tabularnewline
 & {\footnotesize{}KP+SDS} & {\footnotesize{}4.96} & {\footnotesize{}6.58} & {\footnotesize{}10.58} & {\footnotesize{}4.67} & {\footnotesize{}4.97} & {\footnotesize{}5.40} & {\footnotesize{}5.21} & {\footnotesize{}15.08} & {\footnotesize{}12.78} & {\footnotesize{}12.18} & {\footnotesize{}8.24}\tabularnewline
 & {\footnotesize{}PP+SDS} & {\footnotesize{}6.58} & {\footnotesize{}14.02} & {\footnotesize{}14.43} & {\footnotesize{}6.65} & {\footnotesize{}7.96} & {\footnotesize{}7.47} & {\footnotesize{}7.57} & {\footnotesize{}15.21} & {\footnotesize{}15.23} & {\footnotesize{}13.24} & {\footnotesize{}10.84}\tabularnewline
 & {\footnotesize{}Puffball\cite{twarog2012playing}(SDS)} & {\footnotesize{}9.68} & {\footnotesize{}10.23} & {\footnotesize{}11.80} & {\footnotesize{}15.95} & {\footnotesize{}12.42} & {\footnotesize{}8.28} & {\footnotesize{}9.45} & {\footnotesize{}9.60} & {\footnotesize{}23.38} & {\footnotesize{}9.26} & {\footnotesize{}12.00}\tabularnewline
\hline 
\hline 
\multirow{4}{*}{\textbf{\footnotesize{}Depth}} & {\footnotesize{}KP+Mask} & {\footnotesize{}9.02} & {\footnotesize{}7.26} & {\footnotesize{}13.51} & {\footnotesize{}12.10} & {\footnotesize{}8.04} & {\footnotesize{}8.02} & {\footnotesize{}10.00} & {\footnotesize{}23.05} & {\footnotesize{}25.57} & {\footnotesize{}7.48} & {\footnotesize{}12.41}\tabularnewline
 & {\footnotesize{}KP+SDS} & {\footnotesize{}9.07} & {\footnotesize{}7.98} & {\footnotesize{}13.57} & {\footnotesize{}9.90} & {\footnotesize{}7.98} & {\footnotesize{}7.96} & {\footnotesize{}9.99} & {\footnotesize{}22.57} & {\footnotesize{}23.59} & {\footnotesize{}7.64} & {\footnotesize{}12.03}\tabularnewline
 & {\footnotesize{}PP+SDS} & {\footnotesize{}10.94} & {\footnotesize{}11.64} & {\footnotesize{}12.26} & {\footnotesize{}15.95} & {\footnotesize{}13.17} & {\footnotesize{}10.06} & {\footnotesize{}12.55} & {\footnotesize{}21.19} & {\footnotesize{}36.37} & {\footnotesize{}8.98} & {\footnotesize{}15.31}\tabularnewline
 & {\footnotesize{}SIRFS\cite{Barron2012B}} & {\footnotesize{}11.80} & {\footnotesize{}11.83} & {\footnotesize{}15.98} & {\footnotesize{}29.15} & {\footnotesize{}21.64} & {\footnotesize{}15.58} & {\footnotesize{}16.91} & {\footnotesize{}19.64} & {\footnotesize{}37.58} & {\footnotesize{}23.01} & {\footnotesize{}20.31}\tabularnewline
\hline 
\end{tabular}
\caption{\label{tab:ablation} Ablation study for our method assuming/relaxing various annotations at test time on objects in PASCAL VOC. As can be seen, our method degrades gracefully with relaxed annotations. Note that these experiments are in a train/test setting and numbers will differ from table \ref{tab:carvi_compare}. Please see text for more details.}
\end{table*}
Experiments were performed to assess two things: 1) how expressive our learned 3D models are by evaluating how well they matched the underlying 3D shapes of the training data 2) study their sensitivity when fit to images using noisy automatic segmentations and pose predictions. 

\vspace{3mm}
\noindent \textbf{Datasets.} For all our experiments, we consider images from the challenging PASCAL VOC 2012 dataset~\cite{pascal-voc-2012} which contain objects from the 10 rigid object categories (as listed in Table \ref{tab:carvi_compare}). We use the publicly available ground truth class-specific keypoints~\cite{bourdevECCV10} and object segmentations~\cite{BharathICCV2011}. Since ground truth 3D shapes are unavailable for PASCAL VOC and most other detection datasets, we evaluated the expressiveness of our learned 3D models on the next best thing we managed to obtain: the PASCAL3D+ dataset~\cite{pascal3d} which has up to 10 3D CAD models for the rigid categories in PASCAL VOC. PASCAL3D+ provides between 4 different models for ``tvmonitor'' and ``train'' and 10 for ``car'' and ``chair''. The different meshes primarily distinguish between subcategories but may also be redundant (e.g., there are more than 3 meshes for sedans in ``car''). We obtain our subcategory labels on the training data by merging some of these cases, which also helps us in tackling data sparsity for some subcategories. 
The subset of PASCAL we considered after filtering occluded instances, which we do not tackle in this paper, had between 70 images for ``sofa'' and 500 images for classes ``aeroplanes'' and ``cars''. We will make all our image sets available along with our implementation.

\vspace{3mm}
\noindent \textbf{Metrics.} We quantify the quality of our 3D models by comparing against the PASCAL 3D+ models using two metrics - 1) the Hausdorff distance normalized by the 3D bounding box size of the ground truth model \cite{aspert2002mesh} and 2) a depth map error to evaluate the quality of the reconstructed visible object surface, measured as the mean absolute distance between reconstructed and ground truth depth:
\begin{gather}
 Z\text{-MAE}(\hat{Z},Z^{*})=\frac{1}{n\cdot\gamma}\underset{\beta}{\text{min}}\underset{x,y}{\sum}|\hat{Z}_{x,y}-Z^*_{x,y}-\beta|
\end{gather}
where $\hat{Z}$ and $Z^*$ represent predicted and ground truth depth maps respectively. Analytically, $\beta$ can be computed as the median of $\hat{Z}-Z^*$ and $\gamma$ is a normalization factor to account for absolute object size for which we use the bounding box diagonal. Note that our depth map error is translation and scale invariant.
 
 
\subsection{Expressiveness of Learned 3D Models} 


We learn and fit our 3D models on the same whole dataset (no train/test split), following the setup of Vicente et al \cite{carvi14}. Table \ref{tab:carvi_compare} compares our reconstructions on PASCAL VOC with those of this recently proposed method which is specialized for this task (e.g. it is not designed for fitting to noisy data), as well as to a state of the art class-agnostic shape inflation method that reconstructs also from a single silhouette. We demonstrate competitive performance on both benchmarks with our models showing greater robustnes to perspective foreshortening effects on ``trains'' and ``buses''. Category-agnostic methods -- Puffball\cite{twarog2012playing} and SIRFS\cite{Barron2012B} -- consistently perform worse on the benchmark by themselves. Certain classes like ``boat'' and ``tvmonitor'' are especially hard because of large intraclass variance and data sparsity respectively. 

\subsection{Sensitivity Analysis} 

In order to analyze sensitivity of our models to noisy inputs we reconstructed held-out test instances using our models given just ground truth bounding boxes. We compare various versions of our method using ground truth(Mask)/imperfect segmentations(SDS) and keypoints(KP)/our pose predictor(PP) for viewpoint estimation respectively. For pose prediction, we use the CNN-based system of \cite{ShubhamPose} and augment it to predict subtypes at test time. This is achieved by training the system as described in \cite{ShubhamPose} with additional subcategory labels obtained from PASCAL 3D+ as described above. To obtain an approximate segmentation from the bounding box, we use the refinement stage of the state-of-the-art joint detection and segmentation system proposed in \cite{BharathECCV2014}. 

Here, we use a train/test setting where our models are trained on only a subset of the data and used to reconstruct the held out data from bounding boxes. Table \ref{tab:ablation} shows that our results degrade gracefully from the fully annotated to the fully automatic setting. Our method is robust to some mis-segmentation owing to our shape model that prevents shapes from bending unnaturally to explain noisy silhouettes. Our reconstructions degrade slightly with imperfect pose initializations even though our projection parameter optimization deals with it to some extent. With predicted poses, we observe that sometimes even when our reconstructions look plausible, the errors can be high as the metrics are sensitive to bad alignment. The data sparsity issue is especially visible in the case of sofas where in a train/test setting in Table \ref{tab:ablation} the numbers drop significantly with less training data (only 34 instances). Note we do not evaluate our bottom-up component as the PASCAL 3D+ meshes provided do not share the same high frequency shape details as the instance. We will show qualitative results in the next subsection. 

\subsection{Fully Automatic Reconstruction}

\begin{figure*}[htbp!]
  \includegraphics[width = \textwidth]{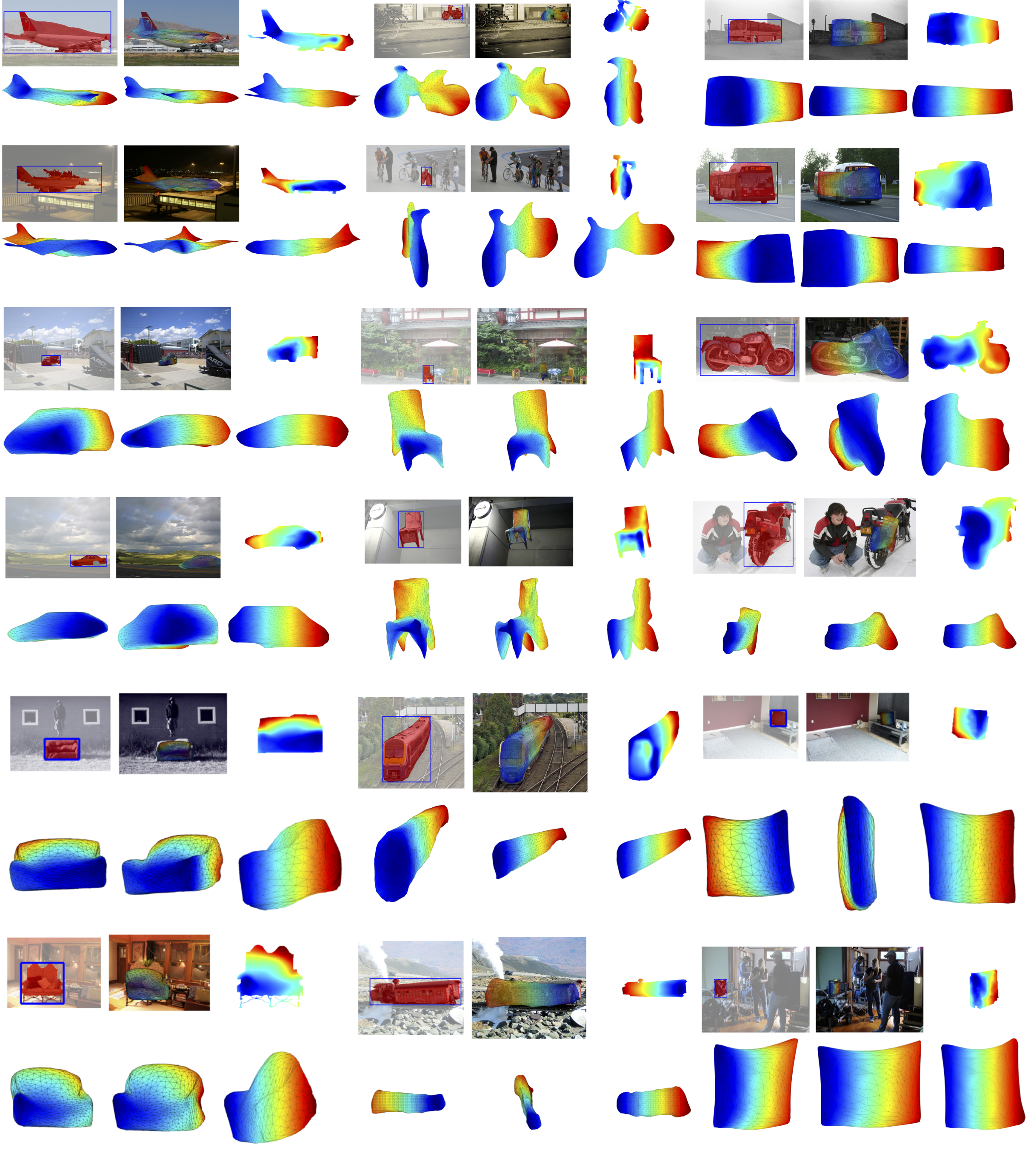}
  \caption{Fully automatic reconstructions on detected instances (0.5 IoU with ground truth) using our models on rigid categories in PASCAL VOC. We show our instance segmentation input, the inferred shape overlaid on the image, a 2.5D depth map (after the bottom-up refinement stage), the mesh in the image viewpoint and two other views. It can be seen that our method produces plausible reconstructions which is a remarkable achievement given just a single image and noisy instance segmentations. Color encodes depth in the image co-ordinate frame (blue is closer). More results can be found at \url{http://goo.gl/lmALxQ}.}
  \label{fig:recons}
\end{figure*}

We qualitatively demonstrate reconstructions on automatically detected and segmented instances with 0.5 IoU overlap with the ground truth in whole images in PASCAL VOC using \cite{BharathECCV2014} in Figure \ref{fig:recons}. We can see that our method is able to deal with some degree of mis-segmentation. Some of our major failure modes include not being able to capture the correct scale and pose of the object and thus badly fitting to the silhouette in some cases. Our subtype prediction also fails on some instances (e.g. CRT vs flat screen ``tvmonitors'') leading to incorrect reconstructions. We include more such images in the supplementary material for the reader to peruse.

\section{Conclusion}
We have proposed what may be the first approach to perform fully automatic object reconstruction from a single image on a large and realistic dataset. Critically, our deformable 3D shape model can be bootstrapped from easily acquired ground-truth 2D annotations thereby bypassing the need for a-priori manual mesh design or 3D scanning and making it possible for convenient use of these types of models on large real-world datasets (e.g. PASCAL VOC). We report an extensive evaluation of the quality of the learned 3D models on a recent 3D benchmarking dataset for PASCAL VOC~\cite{pascal3d} showing competitive results with models that specialize in shape reconstruction from ground truth segmentations inputs while demonstrating that our method is equally capable in the wild, on top of automatic object detectors. 


Much research lies ahead, both in terms of improving the quality and the robustness of reconstruction at test time (both bottom-up and top-down components), developing benchmarks for joint recognition and reconstruction and relaxing the need for annotations during training: all of these constitute interesting and important directions for future work. More expressive non-linear shape models \cite{wu20143d} may prove helpful, as well as a tighter integration between segmentation and reconstruction. 




\section*{Acknowledgements}
This work was supported in part by NSF Award IIS-1212798 and ONR MURI-N00014-10-1-0933. Shubham Tulsiani was supported by the Berkeley fellowship and Jo\~{a}o Carreira was supported by the Portuguese Science Foundation, FCT, under grant SFRH/BPD/84194/2012. We gratefully acknowledge NVIDIA corporation for the donation of Tesla GPUs for this research.

{\small
\bibliographystyle{ieee}
\bibliography{main}
}
\end{document}


\title{Category-Specific Object Reconstruction from a Single Image}
\author{Abhishek Kar$^*$, Shubham Tulsiani$^*$, Jo\~{a}o Carreira and Jitendra Malik\\
University of California, Berkeley - Berkeley, CA 94720\\
{\tt\small \{akar,shubhtuls,carreira,malik\}@eecs.berkeley.edu}}

\maketitle
\blfootnote{* Authors contributed equally}

\section{Additional Results}
We present additional reconstruction results on PASCAL VOC using both the fully automatic and fully annotated settings. As described in the main manuscript, the fully automatic setting involves reconstructing from automatically detected and segmented instances given the full image (using the technique of \cite{BharathECCV2014} in our case). In the fully annotated setting, we present results for the methodology described in section 4.1 where we inspect the expressiveness of our models by reconstructing the whole annotated dataset. 

Figure \ref{fig:supp1} shows results using the fully annotated setting. Note that the reconstructions qualitatively look better owing to poses obtained from ground truth keypoints and clean segmentations. The reconstructed models also tend to fit the silhouettes better. This setting is directly comparable to the class-based reconstruction algorithm presented in \cite{carvi14}. The category ``boat'' is almost always bad because the large intraclass variation results in very noisy estimated cameras in the first place that leads to bad basis shape models being learnt. In some cases like ``trains'', the visual hull initialization of our basis shape models (section 2.2) results in an open front as a result of the isosurface extraction failing. Given this bad initial topology, our algorithm does the best it can to explain the silhouettes in the dataset. Figure \ref{fig:supp2} presents more example reconstructions in the fully automatic setting.

Videos showing more of our results can be found at \url{http://goo.gl/lmALxQ}. The videos show full 3D reconstructions on PASCAL VOC using both settings above and also texture-mapped depth maps with upto 30$^{\circ}$ out of plane rotations.

\newpage

\begin{figure*}[htbp!]
\includegraphics[width = \textwidth]{figures/suppfig1.png}
\caption{Reconstructions obtained using our algorithm on PASCAL VOC using the fully annotated setting (please see text for details). We show the clean segmentation input, the inferred shape overlaid on the image, a 2.5D depth map (after the bottom-up refinement stage), the mesh in the image viewpoint and two other views. Note that here we use ground truth masks and obtain pose using the ground truth keypoints. Color encodes depth in the image co-ordinate frame (blue is closer). More results can be found at \url{http://goo.gl/lmALxQ}}
\label{fig:supp1}
\end{figure*}

\begin{figure*}[htbp!]
\includegraphics[width = \textwidth]{figures/suppfig2.png}
\caption{Fully automatic reconstructions on detected instances (0.5 IoU with ground truth) using our models on rigid categories in PASCAL VOC. We show our instance segmentation input, the inferred shape overlaid on the image, a 2.5D depth map (after the bottom-up refinement stage), the mesh in the image viewpoint and two other views. It can be seen that our method produces plausible reconstructions which is a remarkable achievement given just a single image and noisy instance segmentations. Color encodes depth in the image co-ordinate frame (blue is closer). More results can be found at \url{http://goo.gl/lmALxQ}}
\label{fig:supp2}
\end{figure*}

{\small
\bibliographystyle{ieee}
\bibliography{references}
}